\documentclass[twocolumn]{svjour3}          
\smartqed  
\usepackage{graphicx}
\usepackage{natbib}
\usepackage{amsmath}
\usepackage{amssymb}
\usepackage{enumerate}
\usepackage{lmodern}
\usepackage{multirow}
\usepackage{hhline}
\usepackage{enumitem}

\usepackage{color}
\usepackage{algorithm}
\usepackage{algpseudocode}
\usepackage{booktabs}
\usepackage{subcaption}
\usepackage{url}

\usepackage[colorlinks=true,urlcolor=blue,citecolor=blue,linkcolor=blue,bookmarks=true]{hyperref}

\makeatletter
\newcommand{\UrlAlphabet}{
\do\a\do\b\do\c\do\d\do\e\do\f\do\g\do\h\do\i\do\j%
\do\k\do\l\do\m\do\n\do\o\do\p\do\q\do\r\do\s\do\t%
\do\u\do\v\do\w\do\x\do\y\do\z\do\A\do\B\do\C\do\D%
\do\E\do\F\do\G\do\H\do\I\do\J\do\K\do\L\do\M\do\N%
\do\O\do\P\do\Q\do\R\do\S\do\T\do\U\do\V\do\W\do\X%
\do\Y\do\Z
}
\newcommand{\UrlDigits}{\do\1\do\2\do\3\do\4\do\5\do\6\do\7\do\8\do\9\do\0}
\g@addto@macro{\UrlBreaks}{\UrlOrds}
\g@addto@macro{\UrlBreaks}{\UrlAlphabet}
\g@addto@macro{\UrlBreaks}{\UrlDigits}
\makeatother
\begin{document}

\title{Lightweight and Progressively-Scalable Networks \\for Semantic Segmentation}


\author{Yiheng Zhang \and Ting Yao \and Zhaofan Qiu \and Tao Mei}


\institute{
  Yiheng Zhang \at
  JD Explore Academy, Beijing, China. \\
  \email{yihengzhang.chn@gmail.com}
  \and
  Ting Yao \at
  JD Explore Academy, Beijing, China. \\
  \email{tingyao.ustc@gmail.com}
  \and
  Zhaofan Qiu \at
  JD Explore Academy, Beijing, China. \\
  \email{zhaofanqiu@gmail.com}
  \and
  Tao Mei \at
  JD Explore Academy, Beijing, China. \\
  \email{tmei@live.com}
}

\date{Received: date / Accepted: date}

\maketitle

\hyphenpenalty=5000
\tolerance=1000

\begin{abstract}
  Multi-scale learning frameworks have been regarded as a capable class of models to boost semantic segmentation. The problem nevertheless is not trivial especially for the real-world deployments, which often demand high efficiency in inference latency. In this paper, we thoroughly analyze the design of convolutional blocks (the type of convolutions and the number of channels in convolutions), and the ways of interactions across multiple scales, all from lightweight standpoint for semantic segmentation. With such in-depth comparisons, we conclude three principles, and accordingly devise Lightweight and Progressively-Scalable Networks (LPS-Net) that novelly expands the network complexity in a greedy manner. Technically, LPS-Net first capitalizes on the principles to build a tiny network. Then, LPS-Net progressively scales the tiny network to larger ones by expanding a single dimension (the number of convolutional blocks, the number of channels, or the input resolution) at one time to meet the best speed/accuracy tradeoff. Extensive experiments conducted on three datasets consistently demonstrate the superiority of LPS-Net over several efficient semantic segmentation methods. More remarkably, our LPS-Net achieves 73.4\% mIoU on Cityscapes test set, with the speed of 413.5FPS on an NVIDIA GTX 1080Ti, leading to a performance improvement by 1.5\% and a 65\% speed-up against the state-of-the-art STDC. Code is available at \url{https://github.com/YihengZhang-CV/LPS-Net}.
\end{abstract}

\section{Introduction}\label{sec:intro}
Semantic segmentation is to assign semantic labels to every pixel of an image or a video frame. With the development of deep neural networks, the state-of-the-art networks have successfully pushed the limits of semantic segmentation with remarkable performance improvements. For example, DeepLabV3+ \citep{chen2018encoder} and Hierarchical Multi-Scale Attention \citep{tao2020hierarchical} achieve 82.1\% and 85.4\% mIoU on Cityscapes test set, which are almost saturated on that dataset. The recipe behind these successes originates from multi-scale learning. In the literature, the recent advances involve utilization of multi-scale learning for semantic segmentation along three different dimensions: U-shape~\citep{chen2020fasterseg,Peng_2017_CVPR}, pyramid pooling~\citep{chen2018encoder,zhao2017pspnet}, and multi-path framework~\citep{chen2016attention,tao2020hierarchical}. The U-shape structure hierarchically fuses the features to gradually increase the spatial resolution and naturally produce multi-scale features. The pyramid pooling methods delve into multi-scale information through executing spatial or atrous spatial pyramid pooling at multiple scales. Unlike the former two research schemes, the multi-path frameworks resize the input image to multiple resolutions or scales and feed each scale into an individual path of a deep network. By doing so, the multi-path design places the input resolutions from high to low in parallel and explicitly maintains the high-resolution information rather than recovering from low-scale feature maps. As a result, the learnt features are potentially more capable of classifying and localizing each pixel.

We employ this elegant recipe of multi-path framework and further evolve such type of architectures with good accuracy/speed tradeoff for semantic segmentation. Our philosophies are from two perspectives: (1) lightweighting computational units for semantic segmentation, and (2) progressively scaling up the network while balancing accuracy and inference latency. We propose to explore the first problem by probing the basic unit of convolutional blocks, including the type of convolutions and the number of channels in convolutions, on the basis of several uniqueness (e.g., large feature maps, thin channel widths) in efficient semantic segmentation. Moreover, we further elaborate different ways of interaction across multiple paths with respect to the accuracy/speed tradeoff. Based on these lightweight practice, we build a tiny model, and then progressively expand the tiny model along multiple possible dimensions and select a single dimension that achieves the best tradeoff in each step to alleviate the second issue of accuracy/speed balance.

\begin{figure}
  \centering
  \includegraphics[width=0.95\linewidth]{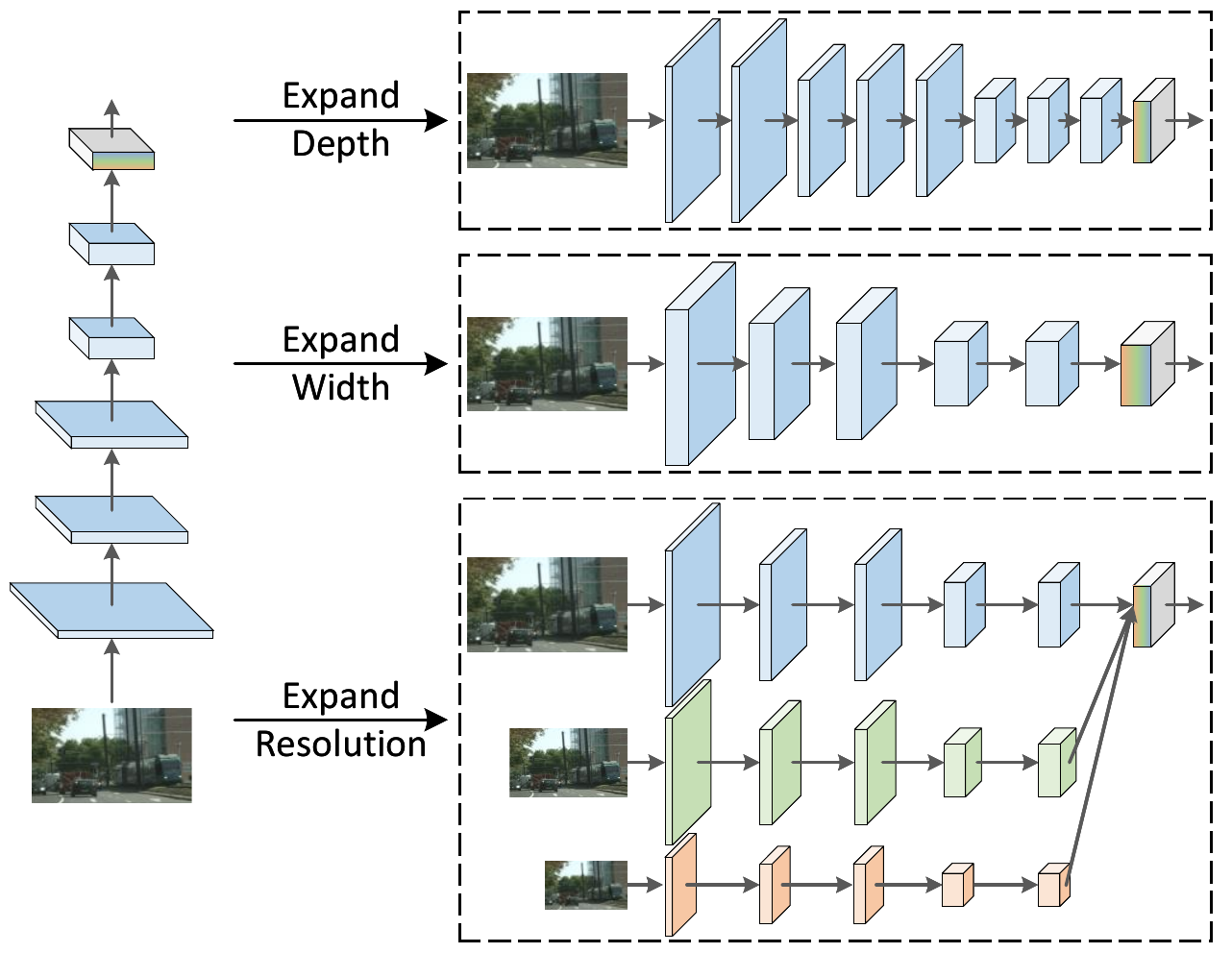}
  \vspace{-0.1in}
  \caption{\small LPS-Net executes the growth of network complexity through expanding a single dimension of the number of convolutional blocks (Depth), the number of channels (Width), or the input resolution (Resolution) at one time.}
  \label{fig:intro_expansion}
  \vspace{-0.1in}
\end{figure}

\begin{figure}
  \centering
  \includegraphics[width=0.95\linewidth]{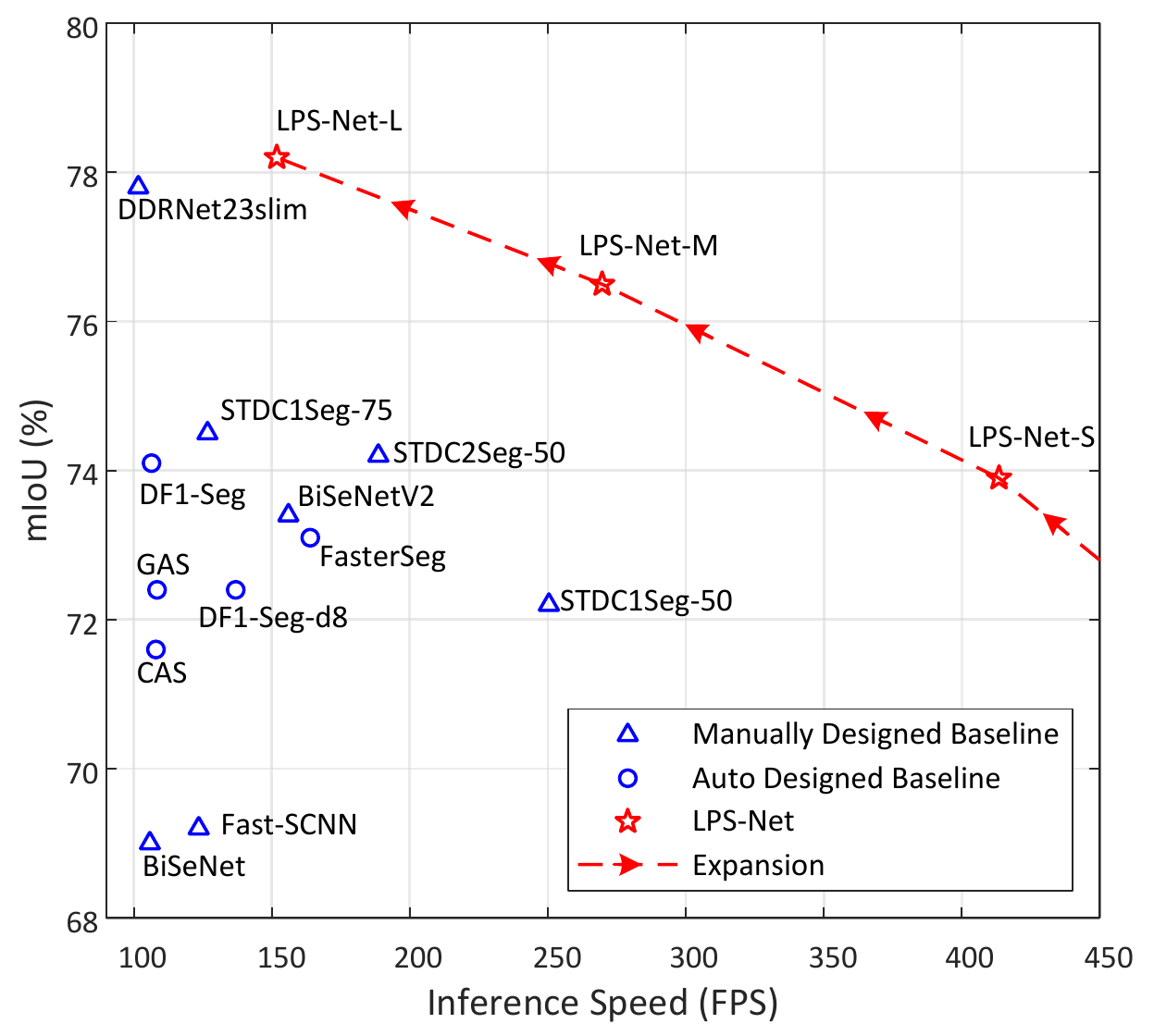}
  \vspace{-0.1in}
  \caption{\small Comparisons of inference speed/accuracy tradeoff on Cityscapes validation set. LPS-Net (-S, -M, and -L) which are progressively expanded along multiple dimensions demonstrate a good balance between accuracy and inference speed compared to other manually-/auto-deigned models.}
  \label{fig:intro_sota}
  \vspace{-0.2in}
\end{figure}

To materialize the idea, we present Lightweight and Progressively-Scalable Networks (LPS-Net) for efficient semantic segmentation. Specifically, LPS-Net bases the multi-path design upon the low-latency regime. Each path takes the resized image as the input to an individual network, which consists of stacked convolutional blocks. The networks across paths share the same structure but have independent learnable weights. The outputs from all the paths are aggregated to produce the score maps, which are upsampled via bilinear upsampling for pixel-level predictions.
In an effort to achieve a lightweight and efficient architecture, we look into the basic unit of convolutional blocks and empirically suggest to purely use $3\times 3$ convolutions with $2^n$-divisible channel widths.
Furthermore, we capitalize on a simple yet effective way of bilinear interpolation to encourage mutual exchange and interactions between paths. With these practical guidelines, LPS-Net first builds a tiny network and then scales the tiny network to a family of larger ones in a progressive manner. Technically, LPS-Net executes the growth of network complexity through expanding a single dimension of the number of convolutional blocks, the number of channels, or the input resolution at one time, as depicted in Figure \ref{fig:intro_expansion}. In our case, LPS-Net ensures a good balance between accuracy and inference speed during expansion, and shows the superiority over other manually-/auto-deigned models, as shown in Figure~\ref{fig:intro_sota}.

In summary, we have made the following contributions: (1) The lightweight design of convolutional blocks and the way of path interactions in multi-path framework are shown capable of regarding as the practical principles for efficient semantic segmentation; (2) The exquisitely devised LPS-Net is shown able to progressively expand the network complexity while striking the right accuracy-efficiency tradeoff; (3) LPS-Net has been properly verified through extensive experiments over three datasets, and superior capability is observed on both NVIDIA GPUs and embedded devices in our experiments.

\section{Related Works}\label{sec:related}
\textbf{Semantic Segmentation.}
With the success of convolutional neural networks (CNNs), FCN~\citep{long2015fully} which deploys CNN in a fully convolutional manner enables dense semantic prediction and end-to-end training for semantic segmentation. Following FCN, researchers propose various techniques, which achieve remarkable performances.
For example, in order to extract multi-scale context information which is known to be critical for pixel labeling tasks~\citep{chen2016attention,ladicky2009associative}, PSPNet~\citep{zhao2017pspnet} applies spatial pyramid pooling and DeepLab~\citep{chen2016deeplab,chen2018encoder} utilizes parallel dilated convolutions with different rates.
\citet{Yuan2021OCNetOC} propose to use object context represented as a dense relation matrix to emphasize the contribution of object information.
Multi-layer feature aggregation~\citep{chen2018encoder,fu2019adaptive,ghiasi2016laplacian,lin2019zigzagnet,Nirkin_2021_CVPR,Peng_2017_CVPR} is performed to recover the spatial information loss caused by spatial reduction in the network. \citet{chen2016attention} and \citet{tao2020hierarchical} capitalize on the multi-path framework to improve the predictions. Such efforts are made to achieve high-quality segmentation but without taking inference latency into account.

\textbf{Lightweight Networks.} The real-world deployments often demand accurate and efficient networks. To accelerate the inference, there have been several techniques, e.g., pruning~\citep{han2015deep}, factorization~\citep{szegedy2016rethinking}, depthwise separable convolution~\citep{chollet2017xception,howard2017mobilenets} and group convolution~\citep{krizhevsky2012imagenet}, being proposed in the literature. To facilitate model deployments and speed up the inference for semantic segmentation, recent works present to manually~\citep{fan2021rethinking,li2019dfanet,li2020semantic,Yu2021BiSeNetVB,Yu_2018_ECCV,Zhao_2018_ECCV} or automatically~\citep{chen2020fasterseg,li2020humans,li2019partial,Lin_2020_CVPR,zhang2019customizable} design lightweight networks. ICNet~\citep{Zhao_2018_ECCV} employs a cascade network structure to achieve real-time segmentation. BiSeNet~\citep{Yu2021BiSeNetVB,Yu_2018_ECCV} treats the spatial details and category semantics of the images separately to obtain a lightweight network.
DFANet~\citep{li2019dabnet} designs a feature reuse method to incorporate multi-level context into encoded features.
SFNet~\citep{li2020semantic} boosts the segmentation by broadcasting high-level features to high-resolution features with cheap operations.
\citet{fan2021rethinking} propose a short-term dense concatenate module to enrich features with scalable receptive field and multi-scale information in a lightweight network.
Lite-HRNet~\citep{Yu2021LiteHRNetAL} utilizes the high-resolution design pattern of HRNet~\citep{Wang2021DeepHR} and introduces conditional channel weighting to replace costly pointwise ($1\times1$) convolutions in shuffle blocks~\citep{zhang2018shufflenet}.
BiAlignNet~\citep{Wu2021FastAA} augments the BiSeNet by employing a gated flow alignment module to align features in a bidirectional way.
To automate lightweight network design, CAS~\citep{zhang2019customizable} presents resource constraints to achieve an accuracy/computation tradeoff and GAS~\citep{Lin_2020_CVPR} further integrates a graph convolution network as a communication mechanism between different blocks.
\citet{li2019partial} prunes the search space with a partial order assumption to search a balanced network.
\citet{chen2020fasterseg} forms a search space integrating multi-resolution branches and calibrates the balance between accuracy and latency by an additional regularization.
AutoRTNet~\citep{Sun2021RealTimeSS} jointly optimizes network depth, downsampling strategies, and the way of feature aggregation to obtain real-time segmentation networks.

Despite having such lightweight networks, the way to balance speed and accuracy for efficient semantic segmentation is not fully explored.
Unlike the works like~\citep{Zhao_2018_ECCV,Yu2021LiteHRNetAL} which model different resolutions with standard Residual Conv blocks, our work aims at probing the basic design of Conv blocks, including the type of convolutions and the number of channels, particularly for efficient semantic segmentation.
Instead of pre-fixing two-path design in networks like BiSeNet~\citep{Yu2021BiSeNetVB,Yu_2018_ECCV} and BiAlignNet~\citep{Wu2021FastAA}, we exquisitely devise a progressive expansion paradigm to scale up the network. Such expansion dynamically balances the depth, width, resolution, and the number of paths of the networks to strike the right accuracy-efficiency tradeoff.

\textbf{Summary.}
Our work focuses on developing light-weight and scalable networks for efficient semantic segmentation upon the low-latency regime. The proposal of LPS-Net contributes by studying not only the practical design of convolutional blocks and the way of path interactions in multi-path framework for semantic segmentation, but also how a scalable multi-path network can be nicely expanded to meet the right accuracy/speed balance.

\begin{figure*}[!tb]
  \centering
  \includegraphics[width=0.99\linewidth]{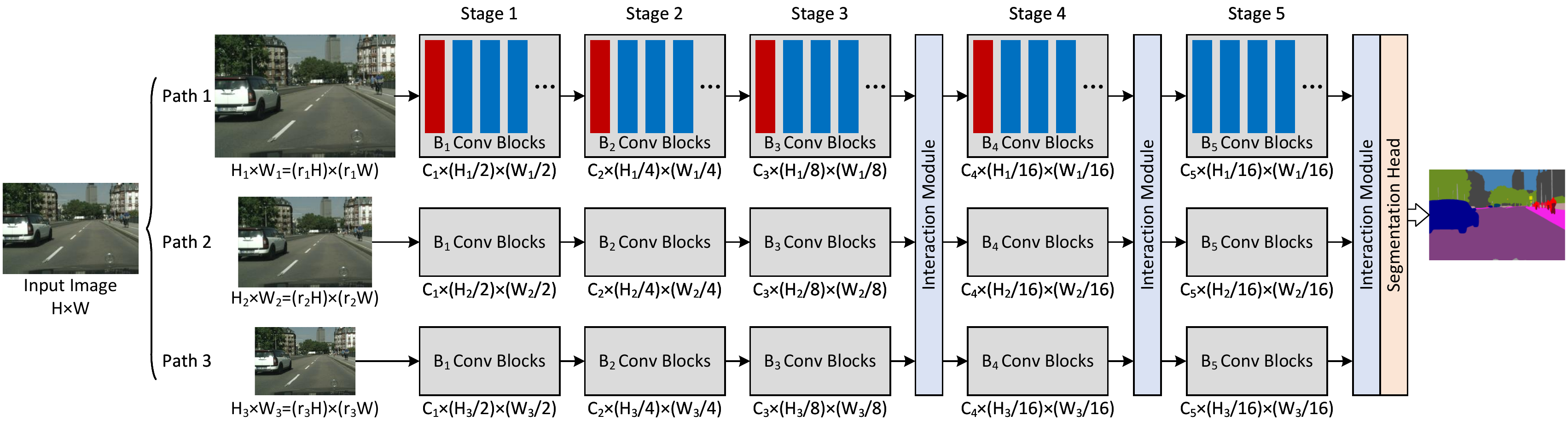}
  \vspace{-0.2in}
  \caption{\small An overview of the design space in our proposed LPS-Net. The \textcolor{red}{red block} and \textcolor{blue}{blue block} represent the convolutional block with stride=2 and stride=1, respectively. The size of output feature map is given for each stage.
  Please note that the 3-path network here is an example to depict the macro architecture, and the number of paths in LPS-Net is dynamically determined by the parameter $\mathcal{R}=\{r_i|^{N}_{i=1}\}$ which is optimized for the right tradeoff along the network expansion.}
  \label{fig:computational_block_stacked}
  \vspace{-0.1in}
\end{figure*}

\section{LPS-Net}
We proceed to present our core proposal, i.e., the Lightweight and Progressively-Scalable Networks (LPS-Net). Specifically, we first introduce the macro architecture in LPS-Net that employs the multi-path recipe. Then, three design principles are presented to upgrade this architecture from lightweight standpoint, including the type of convolutions, the number of channels in convolution, and the way of interaction across multiple paths. Based on these practical guidelines, a family of scalable LPS-Net is devised in a progressive manner by expanding a single dimension at one time to seek the best speed/accuracy tradeoff.

\subsection{Macro Architecture} \label{sec:overall_architecture}
Figure \ref{fig:computational_block_stacked} depicts an overview of the macro architecture in our LPS-Net. The macro architecture is basically constructed in the multi-path design, which resizes the input image to multiple scales and feeds each scale into an individual path. Formally, given the input image with resolution $H\times W$, LPS-Net with $N$ paths (e.g., $N=3$ in Figure \ref{fig:computational_block_stacked}) resizes the image to $r_iH\times r_iW$, which acts as the input of the $i$-th path. Here $r_i$ denotes the scaling ratio. Each path is implemented as a stack of convolutional blocks in five stages, and the $j$-th stage contains $B_j$ blocks. The number of channels (i.e., channel width) are maintained within each stage. Similar to~\citep{chen2018encoder}, the first convolutional block in Stage 1$\sim$4 reduces the spatial dimension by a factor of two. By doing so, the resolution of output feature map from the last stage is $\frac{r_iH}{16}\times \frac{r_iW}{16}$. Moreover, we place interaction modules at the end of Stage 3$\sim$5, aiming to encourage the mutual exchange and interactions between paths. The outputs of all paths are aggregated and fed into a segmentation head to produce the score maps with \textit{num\_classes} channels. Finally, we perform bilinear upsampling over the score maps, yielding the outputs with resolution $H\times W$ that exactly matches the input resolution. Please note that the scaling ratios of paths $\mathcal{R}=\{r_i|^{N}_{i=1}\}$, the repeated number of convolutional blocks $\mathcal{B}=\{B_j|^{5}_{j=1}\}$, and the number of channels $\mathcal{C}=\{C_j|^{5}_{j=1}\}$ in this macro architecture can be flexibly set to make the network structure tailored to the target inference time and adjust the scalability of LPS-Net.

\begin{figure}[!tb]
  \centering
  \includegraphics[width=0.99\linewidth]{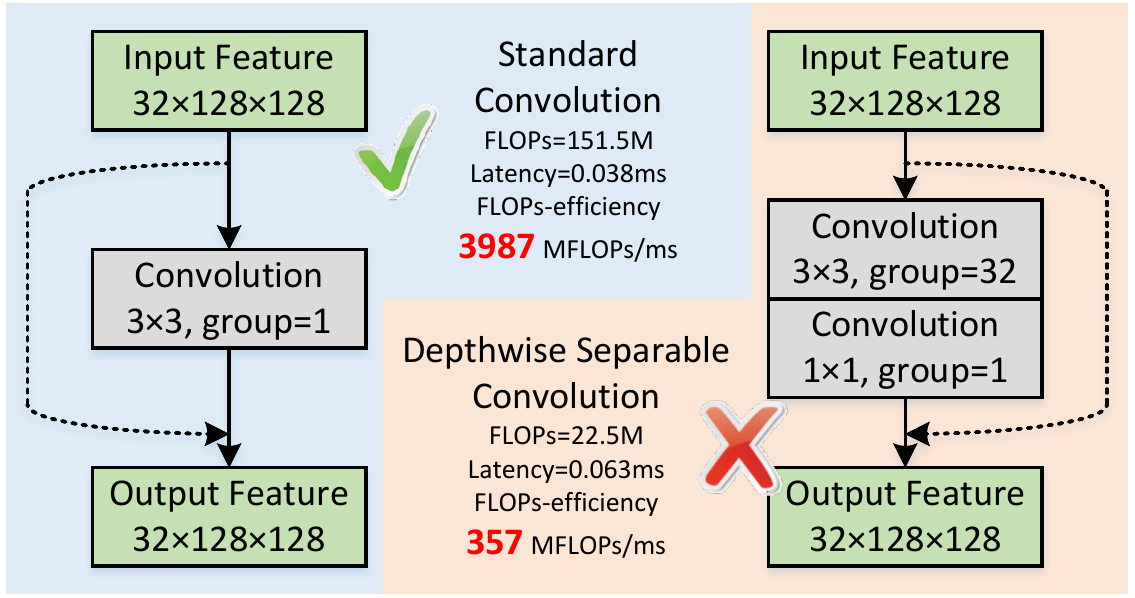}
  \caption{\small A comparison between standard convolution and depthwise separable convolution. The dash line is a optional skip connection and the directed acyclic graph of depthwise separable convolution is a standard version. The shape of input/output feature map is $32\times 128\times 128$, and the activation layers are omitted in the figure.}
  \label{fig:block_comparison_type}
\end{figure}

\subsection{Convolutional Block}\label{ssec:CB}
Convolutional block is the basic computing unit in Convolutional Neural Networks. The stack of convolutional blocks naturally consumes a high percentage of inference time throughout the entire network. Therefore, lightweighting convolutional block is essential for an efficient network. In this section, we delve into the design of basic computational block in LPS-Net along two dimensions: 1) the type of convolutional block, and 2) the number of channels.

\textbf{The Type of Convolutional Block.}
In pursuit of less computational complexity, a series of innovations have been proposed to remould convolutional blocks~\citep{han2020ghostnet,he2016deep,howard2017mobilenets,zhang2018shufflenet}.
In these works, the amount of floating-point operations (FLOPs) is often utilized as the measurement of computational complexity, which guides the design of lightweight network for image recognition. However, FLOPs ignores the memory access cost (MAC) and the degree of parallelism that substantially impact the inference latency of networks, thereby resulting in the discrepancy between FLOPs and actual latency. More importantly, the extension of lightweight designs from image recognition to semantic segmentation is not trivial due to several uniqueness in efficient semantic segmentation (e.g., larger feature maps and thin channel~widths).
To this end, we perform a comparison between the standard convolution and the widely-adopted depthwise separable convolution (SepConv)~\citep{chollet2017xception,howard2017mobilenets,tan2019efficientnet} in Figure~\ref{fig:block_comparison_type}. Here we set the shape of input feature map as $32\times 128\times 128$, which is the most common shape in lightweight networks for semantic segmentation.
The latency is evaluated by executing the block on a PC (i7-8700K/16GB RAM) with an 1070Ti GPU.
The FLOPs-efficiency, which denotes the number of floating-point operations that is processed in per unit of run time, is utilized as the efficiency metric.
As shown in Figure~\ref{fig:block_comparison_type}, the FLOPs-efficiency of standard convolution (3987 MFLOPs/ms) is about 10$\times$ more than that of SepConv (357 MFLOPs/ms).
Accordingly, we utilize the standard convolution as the building block in LPS-Net.

\begin{figure}[!tb]
  \centering
  \includegraphics[width=0.99\linewidth]{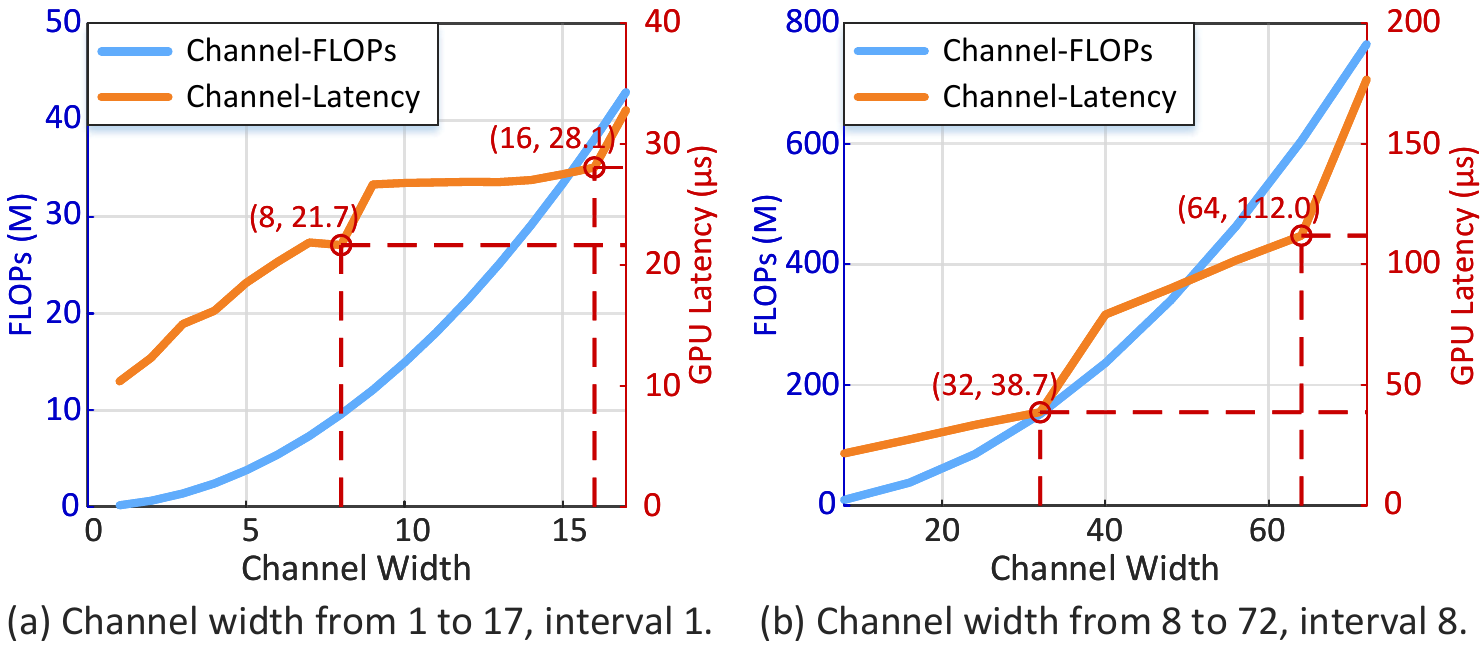}
  \vspace{-0.2in}
  \caption{\small  Efficiency comparisons of $3\times3$ convolutions with channel width from 1 to 17, interval 1 (a), and 8 to 72, interval 8 (b). The spatial resolution of input feature map is fixed as $128\times128$ and the latencies are measured on an NVIDIA GTX 1070Ti GPU with TensorRT.}
  \label{fig:block_comparison_channel}
  \vspace{-0.1in}
\end{figure}

\textbf{The Number of Channels.} The channel width is another important factor that influences the FLOPs-efficiency of convolutions.
Theoretically, the computational complexity necessitates quadratic progression with the increase of channel width.
However, due to the highly-optimized software (e.g., BLAS, cuDNN) and hardware (e.g., CUDA Core, Tensor Core) in modern libraries/devices, the relation between defacto inference time and channel width in practice becomes fuzzy.
To analyze this relation, we compare FLOPs and latency of $3\times 3$ convolutions with different channel widths in Figure~\ref{fig:block_comparison_channel}.
As expected, FLOPs grows quadratically when increasing the channel width from 1 to 17 (see the blue curve in Figure~\ref{fig:block_comparison_channel}(a)).
Meanwhile, the latency only linearly grows and reaches plateaus at around 8 and 16 channel widths (see the red curve in Figure~\ref{fig:block_comparison_channel}(a)).
Such setups of channel width with high FLOPs-efficiency are regarded as the ``sweet spots'' for convolutions.
Similarly, high FLOPs-efficiencies are attained at 32 and 64 channel widths in Figure~\ref{fig:block_comparison_channel}(b).
These sweet points are mainly attributed to the highly parallelizable implementation of convolutions with $2^n$-divisible channel width.
As a result, we make the channel width of the convolution in LPS-Net to be $2^n$-divisible with $n$ as large as possible.

\begin{figure}[!tb]
  \centering
  \includegraphics[width=0.80\linewidth]{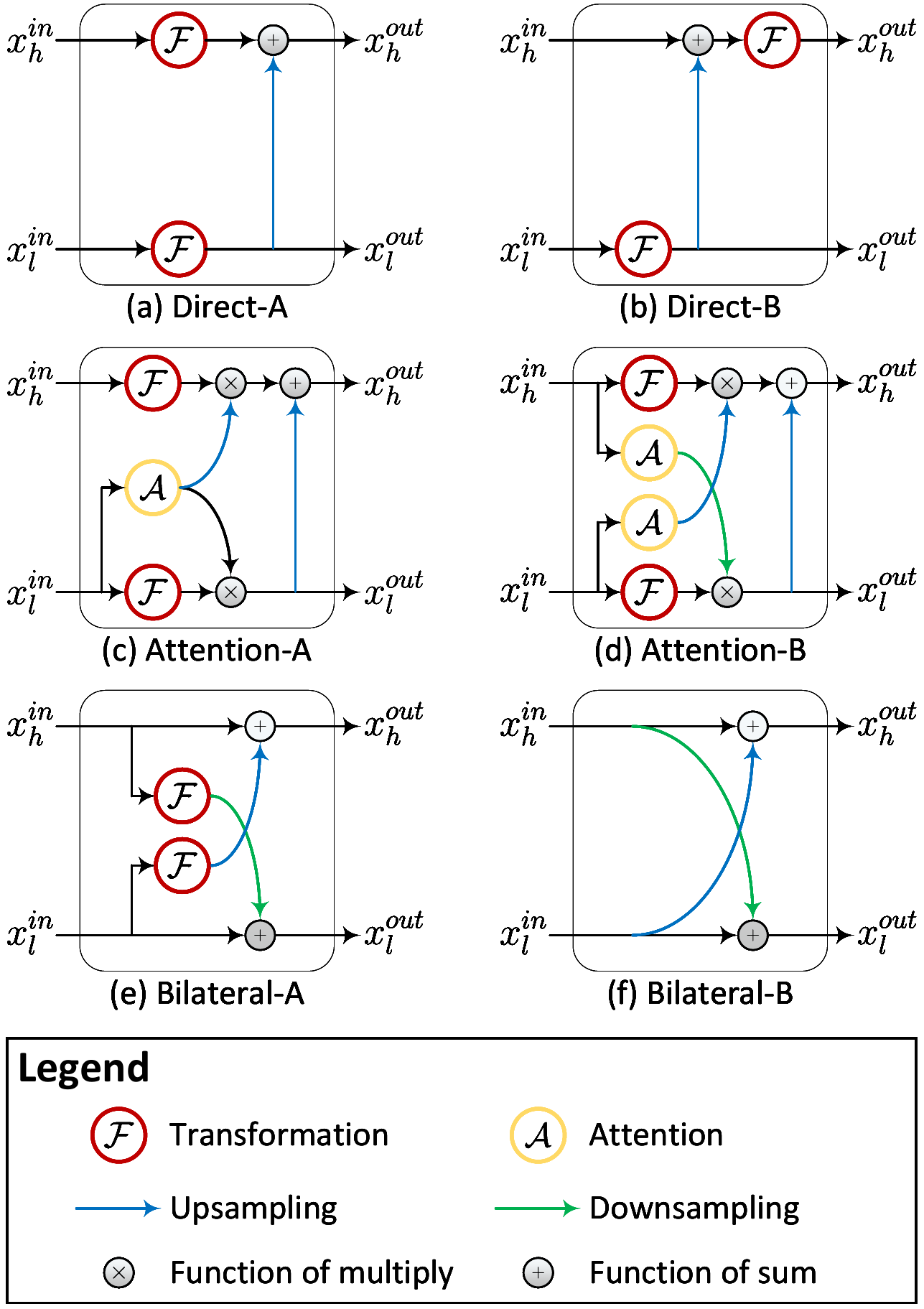}
  \vspace{-0.1in}
  \caption{\small Diagrams of six multi-path interaction modules. These variants of interaction modules can be grouped into three categories: direct feature aggregation (a, b), attention-augmented feature aggregation (c, d), and bilateral feature interaction (e, f).}
  \label{fig:interactions}
\end{figure}

\subsection{Multi-Path Interaction}
In an effort to elegantly diffuse the complementary information across multiple scales, we design the multi-path interaction module that encourages mutual exchange and interactions between pathes in our LPS-Net. Concretely, we define the interaction module $\mathcal{I}$ as $\{x_h^{out}, x_l^{out}\} = \mathcal{I}(\{x_h^{in}, x_l^{in}\})$, where $\{x_h^{in}, x_l^{in}\}$ and $\{x_h^{out}, x_l^{out}\}$ denote the input pair and output pair, respectively. The input pair consists of a high-resolution feature map $x_h^{in}\in\mathbb{R}^{C\times H_h\times W_h}$ and a low-resolution feature map $x_l^{in}\in\mathbb{R}^{C\times H_l\times W_l}$, where $H_h \geq H_l$ and $W_h \geq W_l$. The resolution of each path remains unchanged during the interaction.

Here, we study six variants of interaction modules which can be grouped into three categories (Figure~\ref{fig:interactions}).
For the first category (\textit{Direct-A/B}), the transformed low-resolution features are directly integrated into the high-resolution ones through element-wise summation or channel-wise concatenation. Rather than simply combining the features from two paths, the second category employs attention mechanism to boost the multi-path interaction, namely \textit{Attention-A/B}. Specifically, in \textit{Attention-A}, an attention map $Att(x_l^{in})$ is first calculated from $x_l^{in}$, and we apply it over the transformed $x_h^{in}$ and $x_l^{in}$ before feature aggregation:
\begin{equation}
  \label{eq:interaction_att_a}
  \small
  \left\{
  \begin{aligned}
    x_h^{out} &\!\! =\! \mathcal{U}(\mathcal{F}(x_l^{in})\!*\!Att(x_l^{in}))\!+\!\mathcal{F}(x_h^{in})\!*\!\mathcal{U}(1\!-\!Att(x_l^{in})) \\
    x_l^{out} &\!\! =\! \mathcal{F}(x_l^{in})\!*\!Att(x_l^{in})
  \end{aligned},
  \right.
\end{equation}
where $\mathcal{F}$ and $\mathcal{U}$ denote the convolutional transformation and spatial upsampling, respectively. \textit{Attention-B} further upgrades the simple attention in a bi-direction manner by integrating each path with the attention derived from another path.

In practice, the above four types of interaction modules are computationally expensive due to the heavy transformations and attentions.
Thus, techniques~\citep{chen2016attention,tao2020hierarchical,Yu2021BiSeNetVB,Zhao_2018_ECCV} usually place them at the end of the multi-path networks, while leaving the path interactions at early stages unexploited.
To alleviate this issue, two efficient interaction modules are devised to enable the interactions at early stages, namely \textit{Bilateral-A/B}. \textit{Bilateral-A} transforms low-resolution feature and injects it into the high-resolution one, and vice versa.
Such symmetrical design treats each network path equally and the simple transformation triggers the interaction at early stages in an efficient fashion. \textit{Bilateral-B} further eliminates the convolutional transformation in \textit{Bilateral-A}, leading to the simplest interaction by solely applying spatial resizing and element-wise summation on input feature maps:
\begin{equation}
  \label{eq:interaction_bi_b}
  \small
  \begin{aligned}
    \{x_h^{out}, x_l^{out}\} = \{x_h^{in} + \mathcal{U}(x_l^{in}),~ x_l^{in} + \mathcal{D}(x_h^{in})\}
  \end{aligned}~~,
\end{equation}
where $\mathcal{U}$ and $\mathcal{D}$ denote upsampling and downsampling of the feature maps. The comparisons between all the six variants of interactions will be elaborated in the experiments, and we empirically verify that \textit{Bilateral-B} achieves the best speed/accuracy tradeoff. Therefore, we adopt \textit{Bilateral-B} as the default interaction module in our LPS-Net.

\begin{table*}[!tb]
  \centering
  \caption{\small Expanding operations along three dimensions.}
  \label{tab:scale_op}
  \vspace{-0.1in}
  \begin{tabular}{c|c} \toprule
    Depth $\Delta_{\mathcal{B}}$ & $\{0, 1, 1, 1, 1\},\{0, 0, 1, 1, 1\},\{0, 0, 0, 1, 1\}$ \\
    Width $\Delta_{\mathcal{C}}$ & $\{4, 8, 16, 32, 32\},\{0, 8, 16, 32, 32\},\{0, 0, 16, 32, 32\},\{0, 0, 0, 32, 32\}$ \\
    Resolution $\Delta_{\mathcal{R}}$ & $\begin{matrix} \{1/8, 0, 0\}~\{0, 1/8, 0\}~\{0, 0, 1/8\} \end{matrix}$\\     \bottomrule
  \end{tabular}
\end{table*}

\subsection{Scalable Architectures}
Given the selected convolutional block and interaction module, a particular LPS-Net architecture can be defined as $\mathcal{N}=\{\mathcal{B},\mathcal{C},\mathcal{R}\}$, where $\mathcal{B},\mathcal{C},\mathcal{R}$ are adjustable parameters described in Section~\ref{sec:overall_architecture}. The network complexity is thus determined by these parameters from three dimensions. \textbf{Depth} dimension ($\mathcal{B}$) is the number of blocks which determines the ability of the network to capture the high-level semantic information. \textbf{Width} dimension ($\mathcal{C}$), i.e., the number of channels at each stage, impacts the learning capacity of each convolution. \textbf{Resolution} dimension ($\mathcal{R}$) represents the spatial granularity of each path. To balance the three dimensions when designing LPS-Net architecture, we start by building a tiny network, and then expand a single dimension at one time in a progressive manner.

\begin{algorithm}[!tb]
  \caption{The progressive expansion in LPS-Net.}\label{alg:cap}
  \begin{algorithmic}[1]
  \State \textbf{Input:}
  \State ~~~~Expanding operations $\Delta$
  \State ~~~~number of expansion steps $I$
  \State \textbf{Output:}
  \State ~~~~A family of LPS-Net with different latencies: \\~~~~$\mathbf{N}=\{\mathcal{N}^i|i=0,1,2,3,...,I\}$
  \State \textbf{Initialization:}
  \State ~~~~$\mathcal{N}^0=\{\mathcal{B}^0,\mathcal{C}^0,\mathcal{R}^0\}$
  \State ~~~~~~~~~$=\{\{1,1,1,1,1\},\{4,8,16,32,32\},\{1/2, 0, 0\}\}$
  \State ~~~~$P=Perf(\mathcal{N}^0)$, $L=Lat(\mathcal{N}^0)$, $i=1$
  \State \textbf{Do: }
  \While{$i \leq I$}
  \ForAll{$\delta \in \Delta$}
  \State $L'_\delta = Lat(\mathcal{N}^{i-1} + \delta)$
  \EndFor
  \State $L_T=max(\{L'_\delta|\delta\in\Delta\})$
  \ForAll{$\delta \in \Delta$}
  \State $k_\delta = \mathop{\arg\min} \limits_{k\geq1}|Lat(\mathcal{N}^{i-1}+k\cdot\delta) - L_T|$
  \State $\mathcal{N}_{\delta}^i = \mathcal{N}^{i-1} + k_{\delta}\cdot\delta$
  \State \Comment{\small Expand $\mathcal{N}^{i-1}$ using $\delta$ with stepsize $k_\delta$.}
  \State $P_\delta=Perf(\mathcal{N}_{\delta}^i)$
  \State \Comment{\small Train and evaluate $\mathcal{N}_{\delta}^i$ for semantic segmentation.}
  \State $L_\delta=Lat(\mathcal{N}_{\delta}^i)$
  \EndFor
  \State $\delta^* =\mathop{\arg\max}\limits_{\delta\in\Delta}(P_\delta - P)/(L_\delta-L)$
  \State \Comment{\small Select operation by comparing tradeoffs.}
  \State \textbf{Update:} $\mathcal{N}^i=\mathcal{N}^{i-1} + k_{\delta^*}\cdot\delta^*$, $P=P_{\delta^*}$, $L=L_{\delta^*}$
  \EndWhile
  \end{algorithmic}
\end{algorithm}

Technically, we set the parameters of tiny network $\mathcal{N}^0$ as $\mathcal{B}^0=\{1,1,1,1,1\}$, $\mathcal{C}^0=\{4,8,16,32,32\}$, and $\mathcal{R}^0=\{1/2, 0, 0\}$. Please note that the path with $r=0$ is not included, and thus the tiny network only contains a single path. In this way, the tiny network is very efficient with only 0.38ms latency on a single GPU, and we further scale it to the heavier ones by multi-step expansion. All the candidates of expanding operations $\Delta=\Delta_{\mathcal{B}}\cup\Delta_{\mathcal{C}}\cup\Delta_{\mathcal{R}}$ are defined in Table~\ref{tab:scale_op}.
When $\mathcal{N}=\{\mathcal{B},\mathcal{C},\mathcal{R}\}$ is expanded by $\delta$, $\delta$ will be added to $\mathcal{N}$ along its corresponding dimension and produce the expanded network (e.g. $\{\mathcal{B}+\delta,\mathcal{C},\mathcal{R}\}$ when $\delta\in\Delta_{\mathcal{B}}$).
For each step, we greedily determine the executed expanding operation by comparing the tradeoff between speed and accuracy of each operation $\delta\in\Delta$. Specifically, for the $i$-th step, the target latency after expansion is first determined by the maximum latency across different expanding operations:
\begin{equation}
  \label{eq:calc_lt}
  \small
  \begin{aligned}
    L_T=\max \limits_{\delta\in\Delta}Lat(\mathcal{N}^{i-1} + \delta)
  \end{aligned}~~,
\end{equation}
where $\mathcal{N}^{i-1}$ denotes the architecture from the last step and $Lat(\cdot)$ measures the latency of the network. Then, a stepsize for each operation is calculated as:
\begin{equation}
  \label{eq:calc_k}
  \small
  \begin{aligned}
    k_\delta=\mathop{\arg\min} \limits_{k\geq1}|Lat(\mathcal{N}^{i-1}+k\cdot\delta) - L_T|
  \end{aligned}~~,
\end{equation}
targeting for better aligning the resultant latency to $L_T$ after expanding different operations.
Nevertheless, since the expansion only produces networks in discrete steps, the latencies of the expanded networks may exceed $L_T$. If the selection of operation only depends on the performances of expanded networks, it may be biased to the operations which leads to a larger latency increase. To tackle this issue, the optimal expanding operation $\delta^*$ is finally selected by maximizing the ratio of performance increase and latency increase after expanding at each step:
\begin{equation}
  \label{eq:scaling}
  \small
  \begin{aligned}
    \delta^* &=\mathop{\arg\max}\limits_{\delta\in\Delta}\frac{Perf(\mathcal{N}_{\delta}^i) - Perf(\mathcal{N}^{i-1})}{Lat(\mathcal{N}_{\delta}^{i})-Lat(\mathcal{N}^{i-1})} \\
    \mathcal{N}_{\delta}^i &= \mathcal{N}^{i-1} + k_{\delta}\cdot\delta
  \end{aligned}~~~~,
\end{equation}
where $Perf(\cdot)$ evaluates the performances of semantic segmentation. As such, starting from the tiny network $\mathcal{N}^0$, we iteratively perform the network expansion $I$ times and finally derive a family of LPS-Net with different complexities.
Algorithm~\ref{alg:cap} summarizes the expansion algorithm of LPS-Net.

\section{Experiments}\label{sec:exp}
We empirically verify the merit of our LPS-Net by conducting a thorough evaluation of efficient semantic segmentation on Cityscapes~\citep{Cordts2016Cityscapes}.
Expressly, we first undertake the experiments for semantic segmentation to validate the lightweight designs of the convolutional blocks and the way of interactions across paths in LPS-Net with a pre-fixed structure. The second experiment examines the accuracy and latency tradeoffs along the progressive growth of network complexity of LPS-Net with regard to the number of convolutional blocks, the number of channels, and the input resolution. The third experiment compares three scalable levels of LPS-Net with several state-of-the-art fast semantic segmentation methods and demonstrates the better tradeoffs. Furthermore, we evaluate the transferability of the three versions of LPS-Net, which are learnt on Cityscapes, on BDD100K~\citep{yu2018bdd100k} and CamVid~\citep{brostow2008segmentation} datasets for semantic segmentation.
Finally, we also migrate our LPS-Net to embedded devices and prove the efficacy on two different devices.

\subsection{Datasets}\label{sec:datasets}
The Cityscapes dataset contains 5,000 urban street images with high-quality pixel-level annotations of 19 classes. The image resolution is $1,024\times2,048$. All the images are split into three sets of 2,975, 500 and 1,525 for training, validation, and testing, respectively. For fair comparison, the additional set of 23,473 coarsely annotated images of this dataset is not utilized in our experiments. We train LPS-Net on the training set and evaluate the networks on the validation set. Following the standard protocol in~\citep{fan2021rethinking,Lin_2020_CVPR,zhang2019customizable}, LPS-Net is learnt on the training plus validation sets when submitting the model to online Cityscapes server and reporting the performance on the testing set.

To further examine the generalization of LPS-Net, we re-evaluate the networks on BDD100K~\citep{yu2018bdd100k} and CamVid~\citep{brostow2008segmentation} datasets. BDD100K is a recently released urban street dataset. For semantic segmentation, BDD100K includes 8,000 images being annotated with the consistent labels in Cityscapes. In between, 7,000 and 1,000 images are exploited for training and validation, respectively. The image resolution is $720\times1,280$. CamVid consists of 5 video sequences with resolution $720\times 960$, and is labeled at one frame per second with 11 semantic categories. 468 and 233 labeled frames in CamVid is utilized for training and testing, respectively.

\subsection{Implementation Details}\label{sec:impl}
We implement our proposal on PyTorch~\citep{paszke2019pytorch} platform and employ mini-batch Stochastic Gradient Descent (SGD) algorithm for model optimization. In the \textbf{training stage}, we utilize the cross-entropy loss and train LPS-Net from scratch for 90K iterations, with a batch size of 16, momentum 0.9 and weight decay 0.0005. ``Poly'' policy with power 0.9 is adopted with initial learning rate 0.01. We use a crop size of $768\times768$ and exploit color-jittering, random horizontal flip and random scaling ($0.5\times\sim2.0\times$) to augment training data. Model re-parameterization~\citep{ding2021repvgg} is also adopted for training. To further improve the capability of multi-path networks, we involve more powerful training strategies for the experiments in Section~\ref{sec:comp_sota}, including ImageNet~\citep{Russakovsky2015ImageNetLS} pre-training, large crop size ($768\times1,536$), more training iterations (180K) and online hard element mining~\citep{chen2020fasterseg,fan2021rethinking}. Note that these strategies only enhance the network training and have no influence on model inference.
Such training strategies employed for experiments on Cityscapes dataset are adjusted to train LPS-Net on BDD100K and CamVid datasets.
For the experiments on BDD100K, we use a crop size of $720\times1280$ and the other strategies remain unchanged. For CamVid, we train the LPS-Net for 40K iterations, with a batch size of 12, an initial learning rate of 0.03, and a crop size $720\times960$. ImageNet pre-training and online hard element mining are also involved in experiments on BDD100K and CamVid.

In the \textbf{inference stage}, we input the image into our LPS-Net and perform semantic segmentation without any evaluation tricks (e.g., flipping, sliding-window testing and multi-scale inference), which in general improve the performance but at the expenses of extra latency. The Intersection over Union (IoU) per category and mean IoU over all the semantic categories are utilized as the performance metric.
The inference latency of the network is measured by running networks with batch size of 1 on one NVIDIA GTX 1070Ti (Section~\ref{sec:eval_block} and Section~\ref{sec:eval_scale}),  1080Ti (Section~\ref{sec:comp_sota}) or two embedded devices (Section~\ref{sec:embedded}) with TensorRT in FP32 mode, unless otherwise stated.

\begin{table}[!t]
  \centering
  \setlength{\tabcolsep}{1.6mm}
  \caption{\small Performance comparisons on Cityscapes validation set between multi-path networks which are built on seven types of convolutional block, including ShuffleNet Unit~\citep{ma2018shufflenet}, Inverted Residual~\citep{sandler2018mobilenetv2}, Ghost Module~\citep{han2020ghostnet}, and Residual/Bottleneck~\citep{he2016deep}.}
  \label{tab:exp_block}
    \vspace{-0.05in}
    \begin{tabular}{c|cccc}\toprule
    Convolutional      & mIoU & Latency & FLOPs & FLOPs-eff. \\
    Block              & (\%) & (ms)    & (G)   & (G/ms)     \\\midrule
    ShuffleNet Unit    & 50.8 & 5.11    & 0.8   & 0.15       \\
    Inverted Residual  & 59.2 & 5.12    & 2.0   & 0.39       \\
    Ghost Module       & 58.3 & 5.25    & 2.5   & 0.47       \\
    Residual           & 66.1 & 5.24    & 16.1  & 3.08       \\
    Bottleneck         & 55.7 & 5.65    & 2.2   & 0.39       \\
    $3\times3$ SepConv & 53.9 & 5.10    & 1.1   & 0.21       \\
    $3\times3$ Conv    & \textbf{66.2} & \textbf{5.08}    & 18.5  & 3.63       \\\bottomrule
    \end{tabular}
\end{table}

\begin{table}[!t]
  \centering
  \caption{\small Performance comparisons on Cityscapes validation set between multi-path networks with different interactions.}
  \label{tab:exp_interaction}
    \vspace{-0.05in}
    \begin{tabular}{c|cc|cc}\toprule
      Interaction  & \multicolumn{2}{c|}{mIoU (\%)} & \multicolumn{2}{c}{Latency (ms)} \\\midrule
      Baseline & \multicolumn{2}{c|}{66.9}      & \multicolumn{2}{c}{5.68}         \\\midrule
      Direct-A    & 67.9 & +1.0 & 6.42 & +0.75 \\
      Direct-B    & 68.1 & +1.2 & 6.49 & +0.82 \\
      Attention-A & 68.2 & +1.3 & 6.42 & +0.75 \\
      Attention-B & 68.6 & +1.7 & 6.99 & +1.31 \\
      Bilateral-A & 69.6 & +2.7 & 6.48 & +0.80 \\
      Bilateral-B & \textbf{69.9} & \textbf{+3.0} & \textbf{5.97} & \textbf{+0.29}  \\\bottomrule
    \end{tabular}
\end{table}

\subsection{Evaluation on Lightweight Designs}\label{sec:eval_block}
We first examine the lightweight designs of the convolutional blocks and the ways of interactions across paths in LPS-Net. We start the validation on the impact of the type of convolutional blocks. Specifically, we capitalize on seven convolutional blocks to build the corresponding multi-path networks. Following \citep{chen2016attention,ding2021repvgg}, $\mathcal{B}$ and $\mathcal{R}$ of all the networks are respectively set to $\{1, 2, 4, 7, 7\}$ and $\{1, 1/4, 0\}$. The channel widths $\mathcal{C}$ are set to $\{1c, 2c, 4c, 8c, 8c\}$ for all the networks, where $c$ is to adjust the inference latency of each network to approximate $5ms$ for fair comparisons. In order to purely verify the convolutional blocks irrespective of the interaction influence, we do not include any path interactions here and exploit channel-wise concatenation to aggregate the outputs from all paths of the multi-path network.

\begin{table*}[!tb]
  \centering
  \caption{\small Network Parameters (Depth-$\mathcal{B}$, Width-$\mathcal{C}$ and Resolution-$\mathcal{R}$), latency, mIoU performance and expanding dimension of the network at each expansion step of LPS-Net.}
  \label{tab:exp_networks_params}
  \vspace{-0.1in}
  \begin{tabular}{cccccccc} \toprule
  Step & $\mathcal{B}$                   &$\mathcal{C}$                       & $\mathcal{R}$                    & Latency (ms) & mIoU (\%) & Dimension  & Name \\ \midrule
  0    & \{1, 1, 1, 1, 1\}   & \{4, 8, 16, 32, 32\}    & \{1/2, 0, 0\}    & 0.38         & 24.1      &  -         & -\\
  1    & \{1, 1, 1, 8, 8\}   & \{4, 8, 16, 32, 32\}    & \{1/2, 0, 0\}    & 0.49         & 41.4      & Depth      & -\\
  2    & \{1, 1, 1, 8, 8\}   & \{4, 16, 32, 64, 64\}   & \{1/2, 0, 0\}    & 0.63         & 53.4      & Width      & -\\
  3    & \{1, 1, 1, 8, 8\}   & \{8, 24, 48, 96, 96\}   & \{1/2, 0, 0\}    & 0.98         & 57.8      & Width      & -\\
  4    & \{1, 1, 1, 8, 8\}   & \{8, 24, 48, 96, 96\}   & \{5/8, 0, 0\}    & 1.25         & 60.1      & Resolution & -\\
  5    & \{1, 3, 3, 10, 10\} & \{8, 24, 48, 96, 96\}   & \{5/8, 0, 0\}    & 1.80         & 62.3      & Depth      & -\\
  6    & \{1, 3, 3, 10, 10\} & \{8, 24, 48, 96, 96\}   & \{5/8, 1/4, 0\}  & 2.53         & 64.6      & Resolution & -\\
  7    & \{1, 3, 3, 10, 10\} & \{8, 24, 48, 96, 96\}   & \{3/4, 1/4, 0\}  & 3.37         & 66.1      & Resolution & LPS-Net-S\\
  8    & \{1, 3, 3, 10, 10\} & \{8, 24, 48, 96, 96\}   & \{1, 1/4, 0\}    & 5.17         & 69.5      & Resolution & LPS-Net-M\\
  9    & \{1, 3, 3, 10, 10\} & \{8, 24, 64, 128, 128\} & \{1, 1/4, 0\}    & 7.07         & 70.8      & Width      & -\\
  10   & \{1, 3, 3, 10, 10\} & \{8, 24, 64, 160, 160\} & \{1, 1/4, 0\}    & 9.52         & 71.4      & Width      & LPS-Net-L\\
  11   & \{1, 3, 3, 10, 10\} & \{8, 24, 64, 160, 160\} & \{9/8, 1/4, 0\}  & 12.38        & 72.0      & Resolution & -\\
  12   & \{1, 3, 3, 10, 10\} & \{8, 24, 64, 160, 160\} & \{11/8, 1/4, 0\} & 17.81        & 73.0      & Resolution & -\\
  13   & \{1, 3, 3, 10, 10\} & \{8, 32, 80, 192, 192\} & \{11/8, 1/4, 0\} & 25.14        & 74.2      & Width      & -\\
  14   & \{1, 3, 3, 10, 10\} & \{8, 32, 96, 224, 224\} & \{11/8, 1/4, 0\} & 31.18        & 74.8      & Width      & -\\\bottomrule
  \end{tabular}
\end{table*}

Table~\ref{tab:exp_block} summarizes the mIoU performances, latencies, FLOPs, and FLOPs-efficiencies of the multi-path networks built on seven types of convolutional blocks. Similar to the observations in Section~\ref{ssec:CB}, $3\times3$ SepConv shows a very low FLOPs-efficiency. Conditioning on almost the same latency with $3\times3$ Conv, the utilization of $3\times3$ SepConv results in much fewer FLOPs, making the computational capability of the network weak. $3\times3$ SepConv is also inferior to $3\times3$ Conv in terms of mIoU performance.
Furthermore, ShuffleNet Unit, Inverted Residual and Ghost Module are proven to be three very effective and lightweight convolutional blocks and all three blocks involve $3\times3$ SepConv. Inverted Residual and Ghost Module have benefited from the inverted residual structure with linear bottleneck and a series of linear transformations with cheap cost, respectively. They both greatly improve $3\times3$ SepConv, but the mIoU performances are still lower than that of $3\times3$ Conv. Residual contains a stack of two $3\times3$ convolutions with a shortcut connection and is extended to Bottleneck by stacking three convolutions ($1\times1$, $3\times3$, $1\times1$) instead of two. The use of $1\times1$ convolution in Bottleneck leads to block fragmentation and affects its FLOPs-efficiency. Compared to $3\times3$ Conv, an extra short cut in Residual slightly lowers FLOPs-efficiency and increases the latency by $\sim$3\%.
Benefiting from the high FLOPs-efficiency and better tradeoff, 3$\times$3 Conv exhibits the highest mIoU (66.2\%) with the lowest latency (5.08$ms$). We further analyze the high FLOPs-efficiency of 3$\times$3 Conv from the perspective of devices. FLOPs-lightweight convolutions (e.g. Ghost Module and $3\times3$ SepConv) usually suffer from a low ratio of computing to memory operations~\citep{Gholami2018SqueezeNextHN} and introduce extra overheads like kernel launching and synchronization. In contrast, 3$\times$3 Conv is more friendly to devices and could be accelerated by the Winograd supported in modern computing libraries like cuDNN. As such, 3$\times$3 Conv has high FLOPs-efficiency.
Overall, $3\times3$ Conv achieves the best tradeoff and thus we exploit it as the convolutional block in LPS-Net.

Then, we study how the interactions across paths influence the accuracy-latency tradeoff of the multi-path networks. In this comparison, the network parameters $\mathcal{B}$ and $\mathcal{R}$ are kept unchanged. We fix $c=16$ and exploit $3\times3$ Conv as the convolutional block to construct the multi-path networks. Table \ref{tab:exp_interaction} details the mIoU performance and latency when integrating different path interactions in multi-path networks. As expected, the results across six ways of interactions consistently indicate that using interaction exhibits better mIoU performance and higher latency against Baseline which does not involve path interaction. More specifically, the larger mIoU improvements are attained when employing the Bilateral-A/B interactions than the Direct-A/B and Attention-A/B interactions. Bilateral-B purely basing on bilinear interpolation leads to less increase of latency than Bilateral-A which uses two additional convolutional transformations. Among all the six interactions, Bilateral-B achieves the largest mIoU gain of 3.0\% with the least extra latency of $0.29ms$, and hence we utilize Bilateral-B as the path interaction in our LPS-Net.

\begin{figure}[!tb]
  \centering
  \includegraphics[width=0.85\linewidth]{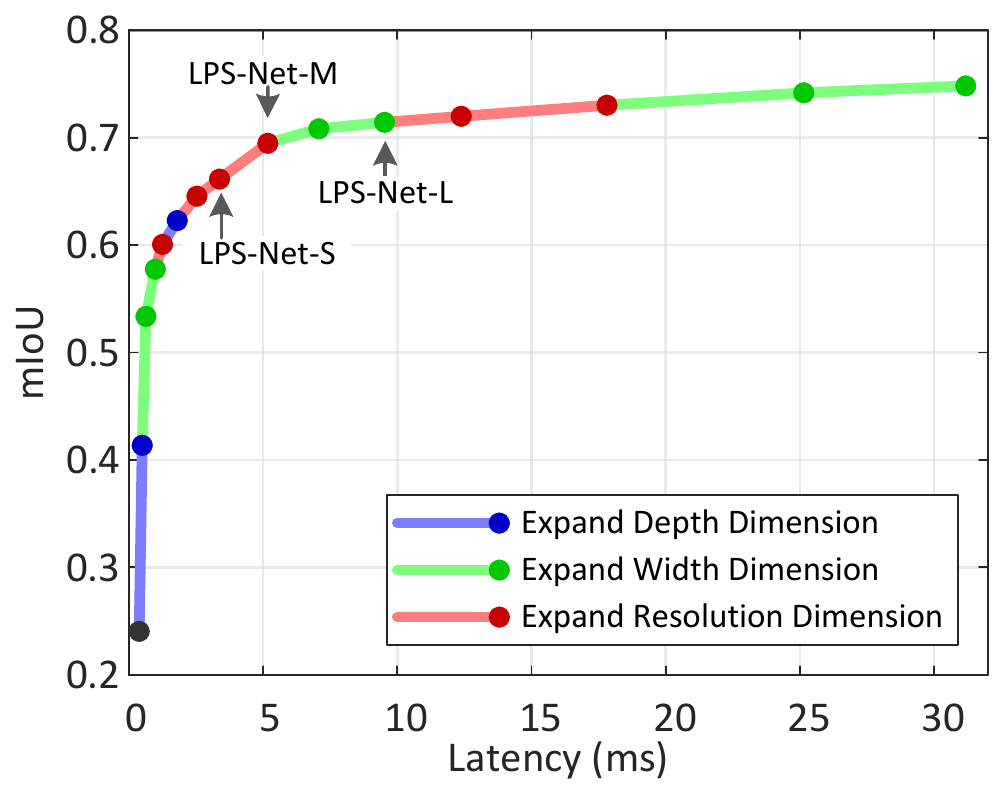}
  \vspace{-0.1in}
  \caption{\small Progressively expanding LPS-Net on Cityscapes. Each dot denotes an expanded networks and each colored segment of the curve represents an expansion along a specific dimension. Note that the inference latencies here are measured with 1070Ti GPU, and performances are evaluated by training networks without powerful strategies described in Section~\ref{sec:impl}.}
  \label{fig:exp_scaling_up}
\end{figure}

\begin{table*}
  \centering
  \captionof{table}{\small Comparisons of mIoU and inference FPS with the state-of-the-art methods on Cityscapes validation \& testing sets.}
  \label{tab:exp_cmp_cityscapes}
  \vspace{-0.1in}
  \begin{tabular}{l|c|cc|c|cc} \toprule
  \multirow{2}{*}{Method}  & \multirow{2}{*}{Resolution} & \multicolumn{2}{c|}{mIoU (\%)} & \multirow{2}{*}{FPS} & \multirow{2}{*}{GPU} & \multirow{2}{*}{Accelerators} \\ \cline{3-4}
                                        &                   & val           & test          &                &          &          \\\midrule
  BiSeNet~\citep{Yu_2018_ECCV}          & $768\times1536$   & 69.0          & 68.4          & 105.8          & TitanXp  &          \\
  BiSeNetV2~\citep{Yu2021BiSeNetVB}       & $512\times1024$   & 73.4          & 72.6          & 156.0          & 1080Ti   & TensorRT \\
  DABNet~\citep{li2019dabnet}           & $512\times1024$   & -             & 70.1          & 104.2          & 1080Ti   &          \\
  DDRNet23slim~\citep{hong2021deep}     & $1024\times2048$  & 77.8          & \textbf{77.4} & 101.6          & 2080Ti   &          \\
  DFANet A~\citep{li2019dfanet}         & $1024\times1024$  & -             & 71.3          & 100.0          & TitanX   &          \\
  DFANet B~\citep{li2019dfanet}         & $1024\times1024$  & -             & 67.1          & 120.0          & TitanX   &          \\
  DFANet A'~\citep{li2019dfanet}        & $512\times1024$   & -             & 70.3          & 160.0          & TitanX   &          \\
  Fast-SCNN~\citep{poudel2019fast}      & $1024\times1024$  & 69.2          & 68.0          & 123.5          & TitanXp  &          \\
  MSFNet~\citep{si2019real}             & $512\times1024$   & -             & 71.3          & 117.0          & 2080Ti   &          \\
  SFNet(DF1)~\citep{li2020semantic}     & $1024\times2048$  & -             & 74.5          & 121.0          & 1080Ti   & TensorRT \\
  STDC1Seg-50~\citep{fan2021rethinking} & $512\times1024$   & 72.2          & 71.9          & 250.4          & 1080Ti   & TensorRT \\
  STDC2Seg-50~\citep{fan2021rethinking} & $512\times1024$   & 74.2          & 73.4          & 188.6          & 1080Ti   & TensorRT \\
  STDC1Seg-75~\citep{fan2021rethinking} & $768\times1536$   & 74.5          & 75.3          & 126.7          & 1080Ti   & TensorRT \\\midrule
  CAS~\citep{zhang2019customizable}     & $768\times1536$   & 71.6          & 70.5          & 108.0          & TitanXp  &          \\
  GAS~\citep{Lin_2020_CVPR}             & $769\times1537$   & 72.4          & 71.8          & 108.4          & TitanXp  &          \\
  AutoRTNet-A~\citep{Sun2021RealTimeSS} & $768\times1536$   & 72.9          & 72.2          & 110.0          & TitanXp  &          \\
  DF1-Seg-d8~\citep{li2019partial}      & $1024\times2048$  & 72.4          & 71.4          & 136.9          & 1080Ti   & TensorRT \\
  DF1-Seg~\citep{li2019partial}         & $1024\times2048$  & 74.1          & 73.0          & 106.4          & 1080Ti   & TensorRT \\
  FasterSeg~\citep{chen2020fasterseg}	  & $1024\times2048$  & 73.1          & 71.5          & 163.9          & 1080Ti   & TensorRT \\
  TinyHMSeg~\citep{li2020humans}        & $768\times1536$   & -             & 71.4          & 172.4          & 1080Ti   & TVM      \\\midrule
  LPS-Net-S                             & $1024\times2048$  & 73.9          & 73.4          & \textbf{413.5} & 1080Ti   & TensorRT \\
  LPS-Net-M                             & $1024\times2048$  & 76.5          & 74.9          & 269.7          & 1080Ti   & TensorRT \\
  LPS-Net-L                             & $1024\times2048$  & \textbf{78.2} & 77.3          & 151.8          & 1080Ti   & TensorRT \\ \bottomrule
  \end{tabular}
\end{table*}

\subsection{Evaluation on Progressively-Scalable Scheme}\label{sec:eval_scale}
Next, we analyze the accuracy-latency balance as the progressive expansion of LPS-Net proceeds. Figure~\ref{fig:exp_scaling_up} depicts the first 14 steps of the process on Cityscapes.
Table~\ref{tab:exp_networks_params} further details the parameters, latency, mIoU performance and expanding dimension of the network at each expansion step of LPS-Net in Figure~\ref{fig:exp_scaling_up}.
We initialize LPS-Net with a tiny structure and the network parameters are $\mathcal{B}^0=\{1,1,1,1,1\}$, $\mathcal{C}^0=\{4,8,16,32,32\}$, $\mathcal{R}^0=\{1/2,0,0\}$. The mIoU and latency of such start point is 24.1\% and $0.38ms$, respectively. As shown in the figure, expanding one certain dimension of Depth (the number of convolutional blocks), Width (the number of channels), or the input Resolution in LPS-Net always improves the mIoU performance.
The fact that the selected dimension is various to strike the best mIoU performance-inference latency tradeoff at different steps necessitates the use of multiple dimensions for expansion.

\begin{figure*}[t]
  \centering\includegraphics[width=0.99\linewidth]{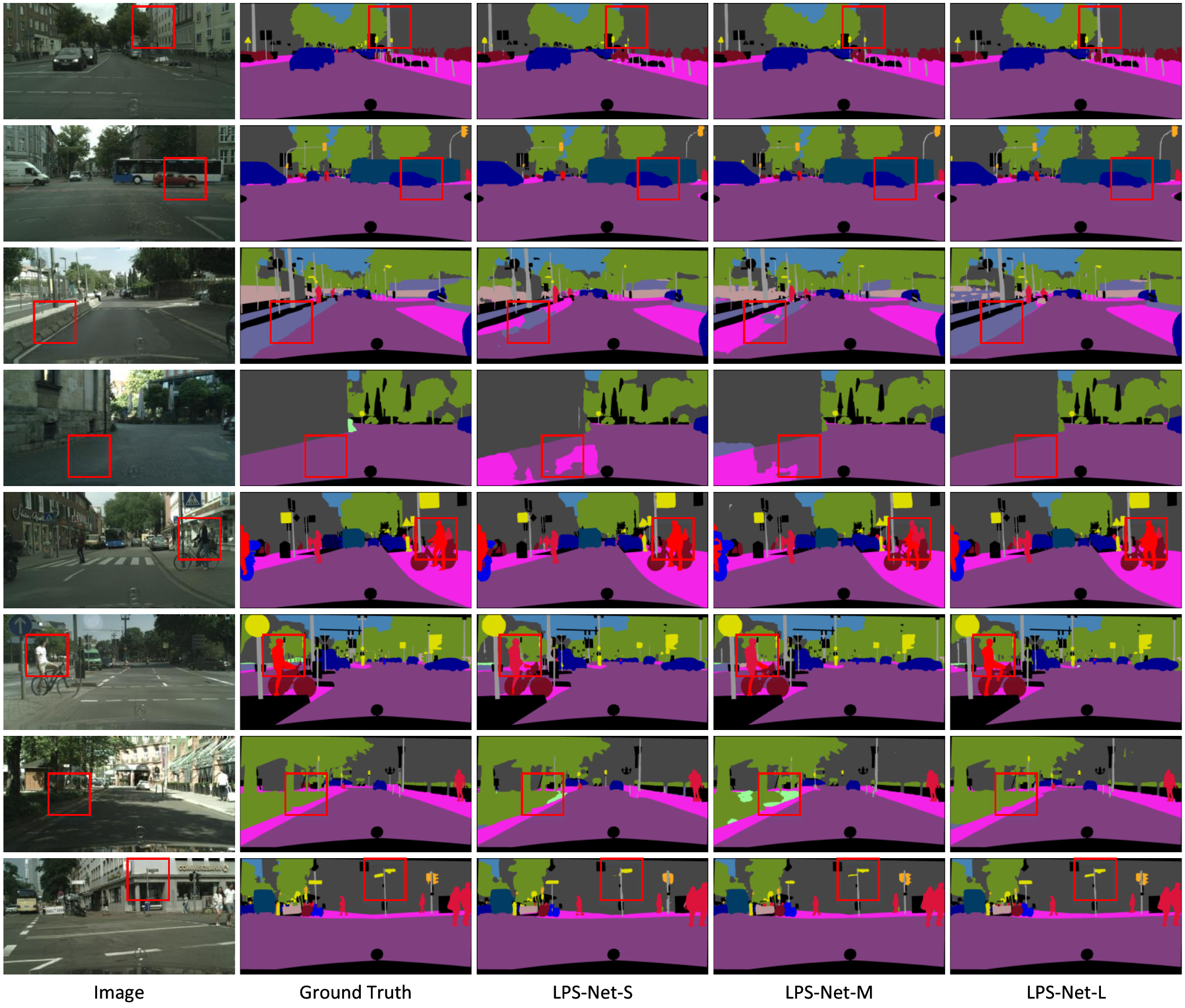}
  \vspace{-0.1in}
  \caption{\small Example of semantic segmentation results in Cityscapes. The improvements could be observed in the red bounding boxes.}
  \label{fig:qualitative}
\end{figure*}

The first step is along the network Depth direction and significantly boosts up the mIoU to 41.4\% with only $0.11ms$ additional latency. We speculate that this may be the result of feature enrichment by stacking more blocks particularly in the context of small channel width and low image resolution. In this case, deepening networks could readily enhance the network capability. The second and third steps further expand Width dimension, and the network after three steps achieves 57.8\% mIoU with less than $1ms$ latency. The results indicate that the network tends to expand the Depth and Width dimensions especially at the very early steps. At the sixth step, the expansion proceeds along the Resolution direction and starts to develop a two-path networks. The next two consecutive steps also work on the factor of Resolution and increase $\mathcal{R}$ to $\{1,1/4,0\}$. The mIoU performance of the two-path networks at the eighth step reaches 69.5\% and the latency is $5.17ms$. That basically verifies the effectiveness of multi-path design to manage the accuracy-latency balance for semantic segmentation. The following steps continue the expansion along either the Width or Resolution dimension.
Such an expansion process indicates some practical insights for efficient semantic segmentation. For example, stacking more blocks for tiny networks could effectively improve the segmentation with negligible extra latency. The multi-path fusion is not always the best choice especially when the networks' inference latency is extremely constrained (e.g., $\leq 2ms$). Compared to increasing channel width and enlarging input resolutions, fusing three or more paths is not optimal for efficient semantic segmentation.

To cover different levels of LPS-Net in complexity, we choose the networks at the seventh, eighth, and tenth steps as the small, medium, and large version of LPS-Net. We name these networks as LPS-Net-S, LPS-Net-M, and LPS-Net-L, respectively.

\subsection{Comparisons with State-of-the-Art Methods}\label{sec:comp_sota}
We compare our LPS-Net with several state-of-the-art fast semantic segmentation methods. To ensure the inference speed is comparable with the most recent works~\citep{chen2020fasterseg,fan2021rethinking,Yu2021BiSeNetVB}, we report the inference FPS of LPS-Net on an NVIDIA GTX 1080Ti GPU card here. Table~\ref{tab:exp_cmp_cityscapes} summarizes the comparisons of mIoU performance and inference FPS on Cityscapes dataset. Note that only the models with FPS$\geq$100 are included in the table. Overall, the results across validation and testing sets consistently indicate that LPS-Net exhibits better tradeoffs than other techniques including hand-crafted networks (e.g.,~\citep{fan2021rethinking,li2019dabnet}) and auto-designed models (e.g.,~\citep{chen2020fasterseg,zhang2019customizable}). In particular, the mIoU score of LPS-Net-S achieves 73.4\% on test set, making the absolute improvement over the best competitor STDC1Seg-50 by 1.5\%. Moreover, LPS-Net-S is faster than STDC1Seg-50 and the inference speed reaches 413.5 FPS, which is impressive. The large version of LPS-Net, i.e., LPS-Net-L, attains 78.2\%/77.3\% mIoU on validation/testing set at 151.8 FPS. It runs 49\% faster than DDRNet23slim with comparable mIoU. This confirms the effectiveness of LPS-Net.
We showcase the qualitative results of LPS-Net-S, LPS-Net-M, and LPS-Net-L in Figure~\ref{fig:qualitative}. Similar to the observation in Table~\ref{tab:exp_cmp_cityscapes}, the quality of prediction is consistently improved when the LPS-Net is incrementally expanded. For example, the prediction in the regions of \textit{pole} (first row), \textit{car} (second row), and \textit{wall} (third row) is amended by expanding the complexity of LPS-Net.

\begin{table}
  \centering
  \captionof{table}{\small Comparisons of mIoU and inference FPS with the state-of-the-art methods on  CamVid testing set. $\dagger$ and $\ddagger$ indicate that the FPS is measured with deep learning accelerators of TensorRT or TVM.}
  \label{tab:exp_cmp_camvid}
  \vspace{-0.05in}
  \begin{tabular}{l|c|c|c} \toprule
  Method                 & mIoU (\%)     & FPS            & GPU      \\ \midrule
  BiSeNet                & 65.6          & 175.0          & TitanXp  \\
  BiSeNet(Res18)         & 68.7          & 116.3          & TitanXp  \\
  BiSeNetV2$^\dagger$    & 72.4          & 124.5          & 1080Ti   \\
  BiSeNetV2-L$^\dagger$  & 73.2          &  32.7          & 1080Ti   \\
  DDRNet23slim           & 74.7          & 230.0          & 2080Ti   \\
  DDRNet23               & 76.3          &  94.0          & 2080Ti   \\
  DFANet A               & 64.7          & 120.0          & TitanX   \\
  DFANet B               & 59.3          & 160.0          & TitanX   \\
  ICNet                  & 67.1          &  27.8          & TitanX   \\
  MSFNet                 & 72.7          & 160.0          & 2080Ti   \\
  MSFNet                 & 75.4          &  91.0          & 2080Ti   \\
  SFNet(DF2)$^\dagger$   & 70.4          & 134.1          & 1080Ti   \\
  SFNet(Res18)$^\dagger$ & 73.8          &  35.5          & 1080Ti   \\
  STDC1Seg-50$^\dagger$  & 73.0          & 197.6          & 1080Ti   \\
  STDC2Seg-50$^\dagger$  & 73.9          & 152.2          & 1080Ti   \\\midrule
  CAS                    & 71.2          & 169.0          & TitanXp  \\
  GAS                    & 72.8          & 153.1          & TitanXp  \\
  AutoRTNet-A            & 73.5          & 140.0          & TitanXp  \\
  FasterSeg$^\dagger$    & 71.1          & 398.1          & 1080Ti   \\
  HMSeg$^\ddagger$       & 75.1          & 130.8          & 1080Ti   \\
  TinyHMSeg$^\ddagger$   & 71.8          & 278.5          & 1080Ti   \\\midrule
  LPS-Net-S$^\dagger$    & 73.6          & \textbf{432.4} & 1080Ti   \\
  LPS-Net-M$^\dagger$    & 75.1          & 317.3          & 1080Ti   \\
  LPS-Net-L$^\dagger$    & \textbf{76.5} & 169.3          & 1080Ti   \\ \bottomrule
  \end{tabular}
\end{table}

To examine the generalization of the network structures of LPS-Net, we further conduct the experiments on CamVid and BDD100K datasets with the three versions of LPS-Net which are learnt on Cityscapes.
Table~\ref{tab:exp_cmp_camvid} details the comparisons of both performance and inference FPS on CamVid.
The first group contains manually designed lightweight models, which are BiSeNet~\citep{Yu_2018_ECCV}, BiSeNetV2~\citep{Yu2021BiSeNetVB}, DDRNet~\citep{hong2021deep}, DFANet~\citep{li2019dfanet}, ICNet~\citep{Zhao_2018_ECCV}, MSFNet~\citep{si2019real}, SFNet~\citep{li2020semantic}, and STDC~\citep{fan2021rethinking}.
The auto-designed lightweight architectures are listed in the second group, including CAS~\citep{zhang2019customizable}, GAS~\citep{Lin_2020_CVPR}, AutoRTNet~\citep{Sun2021RealTimeSS}, FasterSeg~\citep{chen2020fasterseg}, and HMSeg~\citep{li2020humans}.
LPS-Net-S suppresses the fastest competitor FasterSeg by 2.5\% and the inference speed achieves 432.4 FPS. LPS-Net-L yields a better tradeoff (76.5\% mIoU @ 169.3 FPS) than DDRNet23 (76.3\% mIoU@94.0 FPS). Table~\ref{tab:exp_cmp_bdd} shows the comparisons between DRN~\citep{yu2017dilated}, FasterSeg~\citep{chen2020fasterseg}, and LPS-Net on the BDD100K dataset. Our LPS-Net-S again leads to an improvement of mIoU by 0.8\% and a 99\% speed-up against FasterSeg. The results basically validate LPS-Net from the perspective of network generalization.

\subsection{LPS-Net-S on Embedded Devices}\label{sec:embedded}
The real-world scenarios often demand the deployments on embedded devices. Here, we also test the small version of our LPS-Net, i.e., LPS-Net-S, on two devices of a single Nano and a single TX2, both of which are based on NVIDIA Jetson platform.
Table~\ref{tab:exp_cmp_embedded} compares LPS-Net-S with DF1-Seg-d8~\citep{li2019partial}, FasterSeg~\citep{chen2020fasterseg}, BiSeNetV2~\citep{Yu_2018_ECCV}, and STDC1Seg-50~\citep{fan2021rethinking}, regarding the mIoU performance on Cityscapes and inference speed (FPS) on the two devices.
Note that the inference speed is measured by using FP32 data precision to avoid performance degradation. Clearly, LPS-Net-S strikes the superior tradeoff and achieves 10.0/25.6 FPS on Nano/TX2 at resolution $1024\times2048$, which is 41\%/50\% faster than the best competitor STDC1Seg-50~\citep{fan2021rethinking}. The results demonstrate the advantage of LPS-Net on embedded devices.

\begin{table}
  \centering
  \captionof{table}{\small mIoU on BDD100K validation set and inference speed (FPS) on an 1080Ti GPU with TensorRT.}
    \label{tab:exp_cmp_bdd}
    \vspace{-0.05in}
    \begin{tabular}{l|c|c} \toprule
    Method     & mIoU (\%) & FPS   \\ \midrule
    DRN-D-38   & 55.2      & 12.9  \\
    DRN-D-22   & 53.2      & 21.0  \\
    FasterSeg  & 55.1      & 318.0 \\ \midrule
    LPS-Net-S  & 55.9      & \textbf{634.7} \\
    LPS-Net-M  & 58.1      & 495.0 \\
    LPS-Net-L  & \textbf{59.3}      & 283.3 \\ \bottomrule
  \end{tabular}
\end{table}

\begin{table}
  \centering
  \captionof{table}{\small mIoU performance on Cityscapes validation/testing set and inference speed (FPS) on two embedded devices.}
    \label{tab:exp_cmp_embedded}
    \vspace{-0.05in}
    \begin{tabular}{l|c|cc|cc} \toprule
    \multicolumn{1}{l|}{\multirow{2}{*}{Method}} & \multirow{2}{*}{Resolution} & \multicolumn{2}{c|}{mIoU (\%)} & \multicolumn{2}{c}{FPS} \\ \cline{3-6}
    \multicolumn{1}{l|}{}                        &           & val  & test & Nano & TX2  \\ \midrule
    DF1-Seg-d8	       & 1024x2048 & 72.4	& 71.4 & 3.2  &  7.7 \\
    FasterSeg          & 1024x2048 & 73.1 & 71.5 & 4.0  & 10.1 \\
    BiSeNetV2          & 512x1024  & 73.4 & 72.6 & 4.3  & 10.3 \\
    STDC1Seg-50        & 512x1024  & 72.2 & 71.9 & 7.1  & 17.1 \\ \midrule
    LPS-Net-S          & 1024x2048 & \textbf{73.9} & \textbf{73.4} & \textbf{10.0} & \textbf{25.6} \\\bottomrule
  \end{tabular}
\end{table}

\section{Conclusion and Discussion}
We have presented Lightweight and Progressively-Scalable Networks (LPS-Net) which explores the economic design and progressively scales up the network for efficient semantic segmentation. Particularly, we analyze the basic convolutional block and the way of path interaction in multi-path framework which could affect the accuracy-latency tradeoff for semantic segmentation. Empirically, we suggest to use $3\times3$ Conv in convolutional blocks and bilinear interpolation to implement interactions across paths. Based on these guidelines, we first build a tiny network and then extends the tiny network to a series of larger ones through expanding a certain dimension of Width, Depth or Resolution at one time. Experiments conducted on three datasets, i.e., Cityscapes, CamVid and BDD100K, validate our proposal and analysis on both GPUs and embedded devices. More remarkably, our LPS-Net shows impressive accuracy-latency tradeoff on Cityscapes: 413.5 FPS on an NVIDIA GTX 1080Ti GPU or 25.6 FPS on an NVIDIA TX2, with mIoU of 73.4\% on testing set.

\textbf{Data availability.}
The image data that support the findings of this study are available in Cityscapes \citep{Cordts2016Cityscapes} (\url{https://www.cityscapes-dataset.com/}), BDD100K \citep{yu2018bdd100k} (\url{https://bdd-data.berkeley.edu/}) and CamVid \citep{brostow2008segmentation} (\url{http://mi.eng.cam.ac.uk/research/projects/VideoRec/CamVid/}) datasets.

\bibliographystyle{spbasic}      
\bibliography{egbib}   

\end{document}